\documentclass[letterpaper, 10 pt, conference]{ieeeconf}  % Comment this line out if you need a4paper

\IEEEoverridecommandlockouts                              % This command is only needed if 
\usepackage{graphicx}
\usepackage{adjustbox}
\usepackage{booktabs}
\usepackage{amsmath}
\usepackage{amssymb}
\usepackage{url}
\graphicspath{{intro/images/}}
\usepackage{graphicx}
\usepackage{adjustbox}
\usepackage{booktabs}
\usepackage{amsmath}
\usepackage{amssymb}
\usepackage{url}

\usepackage{tabularx}
\usepackage{makecell}
\usepackage{array}

\newcolumntype{Y}{>{\centering\arraybackslash}X}
\title{\LARGE \bf
PACE-Bench: Physical-interaction Attribution and Capability Evaluation for Vision-Language-Action Policies
}

\author{Ta Khanh Ly$^{1}$ and Dung D. Le$^{1}$%
\thanks{$^{1}$Both authors are with the Center for AI Research,
VinUniversity, Hanoi, Vietnam.
{\tt\small lytk2@vinuni.edu.vn},
{\tt\small dungdl@vinuni.edu.vn}}%
}

\begin{document}

\maketitle
\thispagestyle{empty}
\pagestyle{empty}

%%%%%%%%%%%%%%%%%%%%%%%%%%%%%%%%%%%%%%%%%%%%%%%%%%%%%%%%%%%%%%%%%%%%%%%%%%%%%%%%
\begin{abstract}
Task success ranks Vision-Language-Action (VLA) policies, but a single
end-to-end score cannot explain \emph{where} one policy gains an advantage
over another during physical interaction.
We introduce \textbf{PACE-Bench}, an interaction-centric evaluation framework
that maps closed-loop VLA execution into physically meaningful stages:
\textsc{Approach}, \textsc{Contact}, \textsc{Grasp},
\textsc{Stable Hold}, and \textsc{Goal Completion}.
Using evaluator-only simulator state, PACE-Bench supports capability profiling,
cross-model capability comparison, stage-wise failure attribution, and
\textbf{matched performance-gap attribution}, which decomposes an observed
task-success difference into net contributions from stage-specific failure
modes on matched episodes.

We evaluate $\pi_{0.5}$, UniVLA, and OpenVLA on 380 matched episodes per
policy across four LIBERO suites, obtaining task success rates of 97.4\%,
90.8\%, and 71.1\%, respectively.
The attribution analysis reveals that different performance gaps have distinct
interaction-level compositions.
For UniVLA over OpenVLA, 60\% of the net gap is associated with pre-grasp
failures, whereas Place/Insert is the largest classified contributor to the
$\pi_{0.5}$--UniVLA gap (36\%).
Meanwhile, conditional on an observed grasp, \textsc{Stable Hold} is nearly
saturated across all three policies (98.5--99.7\%) despite task-success gaps
as large as 26.3 points.

PACE-Bench therefore turns aggregate benchmark differences into stage-level
behavioral diagnoses, revealing not only whether VLA performance differs,
but which physical failure modes are associated with the observed performance
gap.
\end{abstract}
%%%%%%%%%%%%%%%%%%%%%%%%%%%%%%%%%%%%%%%%%%%%%%%%%%%%%%%%%%%%%%%%%%%%%%%%%%%%%%%%
% ============================================================
\section{Introduction}
\label{sec:introduction}
% ============================================================

Vision-Language-Action (VLA) policies are rapidly improving on
language-conditioned robotic manipulation benchmarks.
Models such as OpenVLA~\cite{kim2024openvla},
$\pi_{0.5}$~\cite{black2025pi05}, and
UniVLA~\cite{wang2025unified} increasingly achieve high success rates on
benchmarks such as LIBERO~\cite{liu2023libero}.
This progress is typically summarized by a single number: task success.

Task success is essential for ranking policies, but it provides only a scalar
view of progress.
If one policy improves from 71\% to 91\% success, the benchmark tells us that
20 percentage points of progress occurred, but not which failure modes are
associated with those additional successes.
This distinction matters for model development.
A manipulation rollout unfolds through physically meaningful interactions:
the robot must approach the relevant object, establish contact, acquire and
maintain a grasp, and finally realize the desired object--target relation.
Different improvements can therefore produce the same increase in aggregate
task success while addressing fundamentally different bottlenecks.

Existing evaluation work has expanded beyond task success through robustness,
safety, execution-quality, and recovery benchmarks.
Failure-recovery methods additionally study behavior after policies deviate
from their intended trajectories~\cite{shin2026b2ff,zhang2026probeact}, while
LIBERO-RECOVER~\cite{liu2026liberorecover} explicitly constructs recovery
scenarios.
These directions reveal important aspects of policy behavior, but they do not
directly answer a different question that becomes increasingly important as
VLA performance improves:

\begin{quote}
\emph{Which physical failure modes are associated with the performance
difference when one VLA policy outperforms another?}
\end{quote}

We introduce PACE-Bench (Physical Interaction Attribution and Capability
Evaluation), an interaction-centric evaluation framework for answering this
question during ordinary closed-loop execution.
PACE-Bench uses privileged simulator state only on the evaluator side to map
rollouts into semantic physical interactions including \textsc{Approach},
\textsc{Contact}, \textsc{Grasp}, \textsc{Stable Hold}, and
\textsc{Goal Completion}.
PACE-Bench provides four complementary views.
An interaction capability profile measures what each policy can reliably
execute.
A cross-model capability comparison identifies where policies differ.
A stage-wise failure attribution assigns unsuccessful rollouts to the first
missing required interaction.
Most importantly, because policies are evaluated on matched episodes,
matched performance-gap attribution decomposes the observed task-success
difference into stage-specific failure contributions on matched episodes.
Thus, PACE-Bench does not merely ask where two policies have different
capability rates; it identifies which stage-specific failures are associated
with the observed leaderboard difference.

We evaluate $\pi_{0.5}$, UniVLA, and OpenVLA on 380 matched episodes per
policy across four LIBERO suites.
Their task-success rates are 97.4\%, 90.8\%, and 71.1\%, respectively.
PACE-Bench reveals that these performance gaps have markedly different
failure-mode compositions.
For UniVLA over OpenVLA, 60\% of the net gap is associated with pre-grasp
failures (Approach, Contact, and Grasp).
In contrast, for $\pi_{0.5}$ over UniVLA, pre-grasp stages account for only
28\% of the net gap, while post-grasp stages account for 40\%;
\textsc{Place/Insert} alone contributes 36\% and is the largest classified
contributor to the remaining gap.
Meanwhile, conditional on an observed grasp, \textsc{Stable Hold} is nearly
saturated across all three policies (98.5--99.7\%), despite task-success
differences as large as 26.3 percentage points.

These results suggest that VLA progress is not uniformly distributed across
physical capabilities.
Large gaps involving the lower-performing policy are primarily associated
with early interaction failures, whereas the smaller gap between the two
stronger policies has its largest classified contribution in downstream
execution.
More broadly, PACE-Bench turns an aggregate benchmark difference into a
stage-level behavioral diagnosis: it can indicate which stages distinguish
policies, which capabilities are already saturated, what type of data may be
most useful to collect next, and whether a proposed method improves the
physical capability it claims to target.

Our contributions are:
\begin{itemize}
    \item We introduce PACE-Bench, an interaction-centric evaluation framework
    that maps ordinary closed-loop VLA rollouts into physically meaningful
    interaction capabilities.

    \item We introduce matched performance-gap attribution, which exploits
    matched episodes to decompose an observed task-success difference into
    stage-specific net failure contributions, going beyond conventional
    comparisons of aggregate or stage-wise success rates.

    \item Through a matched evaluation of $\pi_{0.5}$, UniVLA, and OpenVLA
    across four LIBERO suites, we show that different performance gaps have
    distinct failure-mode compositions: 60\% of the UniVLA--OpenVLA net gap
    is associated with pre-grasp failures, whereas
    \textsc{Place/Insert} is the largest classified contributor to the
    $\pi_{0.5}$--UniVLA gap (36\%).
\end{itemize}
\section{Related Work}
\label{sec:related_work}

% ============================================================
\subsection{Beyond Task Success in VLA Evaluation}
% ============================================================

Standardized benchmarks have played a central role in measuring progress in
language-conditioned robotic manipulation.
CALVIN~\cite{mees2022calvin} evaluates long-horizon language-conditioned
control, LIBERO~\cite{liu2023libero} studies transfer and generalization across
spatial, object, goal, and long-horizon manipulation tasks, and
RoboCasa~\cite{nasiriany2024robocasa} expands evaluation to more diverse
household scenes and manipulation settings.
These benchmarks provide common tasks and protocols for comparing policies,
but performance is still commonly summarized by end-to-end task success.

As VLA policies improve, recent benchmarks increasingly complement task
success with more diagnostic forms of evaluation.
LIBERO-X~\cite{wang2026liberox} evaluates robustness under controlled
distribution shifts in visual appearance, spatial configuration, and language.
Safety-oriented benchmarks such as
SafeVLA-Bench~\cite{fan2026safevlabench} and
LIBERO-Safety~\cite{cui2026liberosafety} separate successful task completion
from safe physical and semantic behavior.
RoboEval~\cite{roboeval2026} further evaluates how a manipulation task is
executed, introducing structured metrics for task progression, coordination,
efficiency, and stability.
Together, these works show that a single success rate can hide important
differences in policy behavior.

PACE-Bench follows this broader shift from ranking policies to diagnosing
their capabilities, but asks a different question.
Rather than introducing new perturbations, safety constraints, or task
variations, we analyze the performance differences that already arise during ordinary matched closed-loop evaluation.
Our goal is not only to determine which policy succeeds more often, but to
identify which physical interaction capabilities are associated with the
observed task-success gap.

This distinction is important because conventional capability metrics and
aggregate task success answer different questions.
A policy may achieve a higher overall success rate without being uniformly
better across all manipulation capabilities, while a capability that appears
important in the task may already be saturated across policies.
PACE-Bench therefore treats aggregate performance improvement as something to
be \emph{decomposed and interpreted}, rather than only reported as a scalar
leaderboard score.

% ============================================================
\subsection{Physical Interaction as an Evaluation Unit}
% ============================================================

Robot manipulation naturally unfolds through a sequence of physically
meaningful interactions.
A typical object-manipulation task requires the robot to approach the relevant
object, establish contact, acquire a grasp, maintain control during transport,
and finally realize the desired object--target relation.
These interactions provide a useful intermediate representation between
low-level robot motion and binary task completion.

Recent work has shown that such intermediate structure is valuable for
reasoning about manipulation progress.
StructVLA~\cite{jin2026structvla}, for example, represents long-horizon
manipulation using sparse, physically meaningful future milestones rather than
dense future observations.
BATON~\cite{xu2026baton} similarly emphasizes transitions between
contact-rich subtasks, where successful completion of one local behavior does
not necessarily guarantee a suitable state for the next.
These works use interaction structure primarily to improve planning or policy
execution.

PACE-Bench instead uses physical interaction as the \emph{unit of evaluation}.
Using evaluator-only simulator state, we derive semantic predicates such as
\textsc{Near}, \textsc{Contact}, \textsc{Holding}, \textsc{On}, and
\textsc{Inside}, and map them to interaction stages including
\textsc{Approach}, \textsc{Contact}, \textsc{Grasp},
\textsc{Stable Hold}, and \textsc{Goal Completion}.

Physical interaction is a particularly useful abstraction for evaluation for
three reasons.
First, it is more interpretable than low-level trajectory error: a deficit in
contact acquisition or goal completion directly corresponds to a recognizable
robot capability.
Second, interaction stages recur across many manipulation tasks and are
therefore less task-specific than manually defined subgoals.
Third, their ordered physical structure provides a common representation for
comparing policies at multiple points along the manipulation process.

PACE-Bench combines this representation with matched-episode evaluation.
Stage-wise capability comparison reveals where two policies differ, while
matched performance-gap attribution goes further by decomposing the observed
task-success difference into stage-specific net contributions.
Thus, physical interaction serves not only as a descriptive representation of
robot behavior, but as the level at which aggregate VLA progress can be
interpreted and compared.
% ============================================================
\section{PACE-Bench}
\label{sec:PACEbench}
% ============================================================

\begin{figure*}[t]
    \centering
    \includegraphics[width=\textwidth]{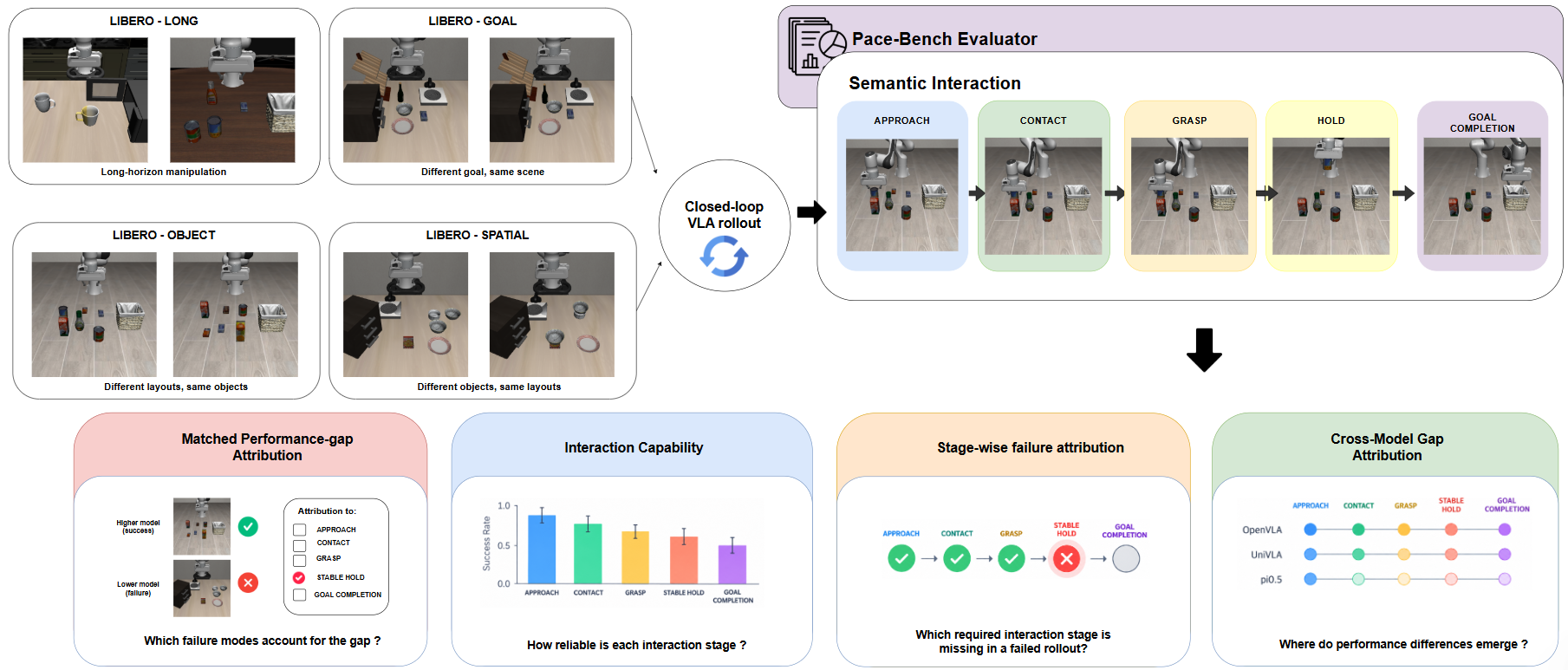}
    \caption{
        \textbf{Overview of PACE-Bench.}
        Given standard closed-loop rollouts, PACE-Bench uses evaluator-only
        simulator state to identify semantic physical interaction stages and
        provides four complementary analyses: interaction capability profiling,
        cross-model capability comparison, matched performance-gap attribution,
        and stage-wise failure attribution.
    }
    \label{fig:ove}
\end{figure*}

PACE-Bench (\textbf{P}hysical Interaction \textbf{A}ttribution and
\textbf{C}apability \textbf{E}valuation) is an interaction-centric evaluator
for closed-loop robotic manipulation.
As illustrated in Fig.~\ref{fig:ove}, PACE-Bench maps ordinary closed-loop
rollouts into physically meaningful interaction stages and provides four
complementary views:
(i) interaction capability profiles,
(ii) cross-model capability comparison,
(iii) matched performance-gap attribution, and
(iv) stage-wise failure attribution.

% ============================================================
\subsection{Semantic Interaction Representation}
\label{sec:PACE_interactions}
% ============================================================

For episode $i$, let
\[
\tau_i=\{(o_t,a_t,s_t)\}_{t=1}^{T_i}
\]
denote the closed-loop rollout, where $o_t$ and $a_t$ are the policy
observation and executed action, and $s_t$ is simulator state available only
to the evaluator.

From $s_t$, PACE-Bench extracts five binary predicates describing the physical
interaction between the robot, object, and task target:
\textsc{Near}, \textsc{Contact}, \textsc{Holding}, \textsc{On}, and
\textsc{Inside}.
These predicates are obtained from MuJoCo geometry and contact information,
robosuite grasp checks, and LIBERO object--target relations.

Simulator signals may briefly flicker because of transient contacts or
geometric noise.
We therefore temporally debounce each predicate.
A predicate is activated only after $k_{\mathrm{on}}$ consecutive positive
observations and deactivated only after $k_{\mathrm{off}}$ consecutive
negative observations.
We use $(k_{\mathrm{on}},k_{\mathrm{off}})=(2,3)$ for
\textsc{Near}/\textsc{Contact} and $(3,5)$ for
\textsc{Holding}/\textsc{On}/\textsc{Inside}.

The filtered predicates define five primary interaction stages in canonical
order: \textsc{Approach}, \textsc{Contact}, \textsc{Grasp},
\textsc{Stable Hold}, and \textsc{Goal Completion}.
\textsc{Approach} is reached when the robot is stably near the relevant object,
and \textsc{Contact} when stable physical contact is established.
\textsc{Grasp} is successful when a valid \textsc{Holding} state is acquired
within 40 timesteps after \textsc{Contact}.
\textsc{Stable Hold} is evaluated once a grasp event has occurred, including
grasps acquired outside the 40-step \textsc{Grasp} window, and requires
\textsc{Holding} to persist for at least 10 timesteps.
\textsc{Goal Completion} is reached when the task-relevant
\textsc{On} or \textsc{Inside} relation is satisfied.

Because \textsc{Grasp} measures timely acquisition whereas
\textsc{Stable Hold} is conditioned on an observed grasp event, the capability
dimensions are diagnostic stage metrics rather than strictly nested event sets.
A late grasp may therefore fail the \textsc{Grasp} metric while remaining
eligible for \textsc{Stable Hold}.

Release can also be detected from the transition out of
\textsc{Holding}, but it is excluded from the primary capability profile
because LIBERO often terminates once the task goal is satisfied, making
release only partially observable.

% ============================================================
\subsection{Interaction Capability Profile}
\label{sec:PACE_capability}
% ============================================================

PACE-Bench first measures how reliably each policy executes each interaction
stage.
For stage $k$, let $\mathcal{E}_k$ denote the set of eligible records and let
$c_{j,k}\in\{0,1\}$ indicate whether record $j$ successfully realizes that
stage.
The capability score is

\begin{equation}
C_k =
\frac{\sum_{j\in\mathcal{E}_k} c_{j,k}}
     {|\mathcal{E}_k|}.
\label{eq:capability}
\end{equation}

Thus, $C_k$ is the fraction of valid records in which stage $k$ is completed
successfully.
Eligibility is stage-specific: grasp-optional tasks and records with unreliable
holding signals are excluded from holding-dependent metrics, while censored
records are removed from the corresponding denominator rather than counted as
failures.

For each eligible record $j$ with an observed grasp event, let
$y_j^{\mathrm{goal}}\in\{0,1\}$ indicate whether the task-relevant goal
relation is satisfied at the nominal endpoint.
We define

\begin{equation}
\mathrm{GCG}_{\mathrm{end}}
=
\frac{1}{|\mathcal{G}|}
\sum_{j\in\mathcal{G}} y_j^{\mathrm{goal}},
\label{eq:gcg}
\end{equation}

where $\mathcal{G}$ is the set of eligible records with an observed grasp
event.
Unlike the stage-transition metrics, GCG$_{\mathrm{end}}$ is evaluated at the
nominal endpoint and is therefore not subject to transition-boundary censoring.

The resulting capability profile is
\[
[
C_{\mathrm{app}},
C_{\mathrm{con}},
C_{\mathrm{grasp}},
C_{\mathrm{hold}},
\mathrm{GCG}_{\mathrm{end}}
].
\]

This profile characterizes what each policy can reliably execute, rather than
only whether the full task eventually succeeds.

% ============================================================
\subsection{Cross-Model Capability Comparison}
\label{sec:PACE_compare}
% ============================================================

To identify where two policies differ, we compare their capability scores
stage by stage:

\begin{equation}
\Delta_k(\pi_A,\pi_B)
=
C_k(\pi_A)-C_k(\pi_B).
\label{eq:capability_gap}
\end{equation}

A positive $\Delta_k$ means that $\pi_A$ has a higher stage-specific
capability rate than $\pi_B$ under the corresponding eligibility definition.
These differences reveal whether policies differ broadly across the interaction
chain or mainly at particular stages.

However, they do not decompose the task-success gap, because capability scores
are computed over stage-specific eligible records.
Moreover, these eligible sets may differ across policies, so the resulting
cross-model differences should be interpreted as comparisons of conditional
stage rates rather than unconditional capability differences over a common
task-defined population.
We therefore separate capability comparison from the matched performance-gap
attribution defined next.

% ============================================================
\subsection{Matched Performance-Gap Attribution}
\label{sec:PACE_matched_gap}
% ============================================================

Because two policies are evaluated on the same tasks and initial states,
PACE-Bench can directly decompose their observed task-success difference.

Assume $\pi_A$ has higher overall task success than $\pi_B$.
Let $\mathcal{W}$ contain episodes where $\pi_A$ succeeds and $\pi_B$ fails,
and let $\mathcal{L}$ contain episodes where $\pi_B$ succeeds and $\pi_A$
fails.
Episodes where both policies have the same outcome do not affect their
success-rate difference.
Therefore,

\begin{equation}
\Delta_{\mathrm{succ}}(\pi_A,\pi_B)
=
\frac{|\mathcal{W}|-|\mathcal{L}|}{N},
\label{eq:success_gap}
\end{equation}

where $N$ is the number of matched episodes.
We refer to $|\mathcal{W}|-|\mathcal{L}|$ as the
\emph{net success gain}: the number of matched episodes by which the
higher-performing policy has a net advantage over the lower-performing one.
Dividing this quantity by $N$ gives the task-success gap.

We then group these discordant episodes by the failure stage of the
unsuccessful policy.
Let $\mathcal{W}_k$ and $\mathcal{L}_k$ denote the corresponding subsets for
stage $k$.
The net contribution of stage $k$ to the task-success gap is

\begin{equation}
A_k(\pi_A,\pi_B)
=
\frac{|\mathcal{W}_k|-|\mathcal{L}_k|}{N}.
\label{eq:stage_attribution_pp}
\end{equation}

We additionally report the fraction of the total net gap associated with
stage $k$:

\begin{equation}
R_k(\pi_A,\pi_B)
=
\frac{|\mathcal{W}_k|-|\mathcal{L}_k|}
     {|\mathcal{W}|-|\mathcal{L}|}.
\label{eq:stage_attribution_share}
\end{equation}

$R_k$ is a signed normalized contribution to the net success gap.
Therefore, unlike an ordinary composition percentage, $R_k$ may be negative
or exceed 100\% when positive and negative stage contributions offset one
another.
It is defined and reported only when
$|\mathcal{W}|-|\mathcal{L}|\neq0$.

When every discordant episode receives a valid label,
\[
\sum_k A_k=\Delta_{\mathrm{succ}},
\qquad
\sum_k R_k=1.
\]

Unclassifiable episodes are retained as a separate category.
Thus, matched attribution identifies which stage-specific failure modes are
associated with the observed task-success difference between two policies.

% ============================================================
\subsection{Stage-Wise Failure Attribution}
\label{sec:PACE_failure}
% ============================================================

Each unsuccessful rollout is assigned a semantic failure stage.
PACE-Bench selects the first required interaction missing from the canonical
task sequence:

\begin{equation}
F_i =
\min_{\prec}
\left\{
k\in\mathcal{S}_i :
k \text{ is required but not observed}
\right\},
\label{eq:failure}
\end{equation}

where $\prec$ denotes semantic stage order rather than chronological failure
time.

For failure analysis, we use the finer sequence
\textsc{Approach}, \textsc{Contact}, \textsc{Grasp},
\textsc{Retain}, \textsc{Release}, and \textsc{Place/Insert}.
\textsc{Retain} denotes failure to maintain the grasp sufficiently for
continued execution and is reported as \textsc{Stable Hold}.
Stages not required by a task are skipped.

For multi-object tasks, objects are evaluated in the order specified by the
task goal, and the stages above are checked in canonical order within each
object.
The episode is assigned the first missing required stage of the first object
with an incomplete interaction sequence.
Episodes that cannot be assigned reliably, including failed episodes for which
no required stage is missing, are labeled \emph{Unclassifiable}.

This attribution is behavioral rather than causal: it identifies the first
required interaction missing from the semantic task progression, not the exact
mechanism or chronological instant of failure.

% ============================================================
\subsection{Evaluation Protocol}
\label{sec:PACE_protocol}
% ============================================================

All metrics are computed within the nominal rollout window, ending at
environment \texttt{done} for successful episodes or at the benchmark step
limit otherwise.
Stored post-\texttt{done} trajectory tails are excluded.

Because temporal debouncing requires future observations, events occurring near
the rollout boundary may be censored.
Such records are excluded from the relevant capability denominator rather than
counted as successes or failures.

All policies retain their native observation processing, action representation,
chunking, replanning strategy, and inference procedure.
Comparability is enforced through matched tasks and initial states together
with a shared PACE-Bench evaluator and scoring protocol.
% ============================================================
\section{Experiments and Results}
\label{sec:experiments}
% ============================================================

% ============================================================
\subsection{Experimental Setup}
\label{sec:experimental_setup}
% ============================================================

\noindent\textbf{Benchmark.}
We evaluate PACE-Bench on LIBERO-Spatial, LIBERO-Object, LIBERO-Goal,
and LIBERO-10 (Long-Horizon)~\cite{liu2023libero}.
All policies are evaluated on the same fixed episode manifest and matched
initial conditions.
The original evaluation contains 400 episodes per policy.
After excluding 20 unary-goal-only episodes unsupported by the current
interaction evaluator, the primary analysis contains 380 matched episodes and
440 manipulated-object channels per policy.
The excluded set consists of two unary-goal LIBERO-Goal tasks with 10 trials
each and is identical across all three policies.
Episodes involving multiple manipulated objects contribute multiple
object-level records, yielding 440 manipulated-object channels.

\noindent\textbf{Evaluated policies.}
We evaluate three representative VLA policies:
UniVLA~\cite{wang2025unified},
$\pi_{0.5}$~\cite{black2025pi05},
and OpenVLA-7B~\cite{kim2024openvla}.
Each policy retains its native observation processing, action representation,
chunking, replanning strategy, and inference procedure.
No privileged simulator state is provided to the evaluated policies.

\noindent\textbf{Statistical analysis.}
For Approach, Contact, Grasp, and Stable Hold, 95\% confidence intervals are
estimated using episode-cluster bootstrap resampling with 10,000 samples.
For GCG$_{\mathrm{end}}$, we report 95\% Wilson intervals.
Cross-policy capability differences are tested using paired permutation tests
on matched episodes, with Holm correction within each policy pair.
For matched performance-gap attribution, 95\% confidence intervals are
estimated using paired episode bootstrap with 10,000 resamples.
Statistical significance is assessed at $\alpha=0.05$.

% ============================================================
\subsection{Task-Level Performance}
\label{sec:task_success}

We first report conventional task success in
Table~\ref{tab:task_success_suite}, which provides the aggregate ranking
that PACE-Bench subsequently analyzes.

\begin{table}[t]
    \centering
    \caption{
        Task success across LIBERO suites.
        Values are successful episodes / total episodes, with success rate
        (\%) in parentheses.
    }
    \label{tab:task_success_suite}

    \setlength{\tabcolsep}{2.6pt}
    \renewcommand{\arraystretch}{1.08}
    \footnotesize

    \begin{tabular}{lccc}
        \toprule
        \textbf{Suite}
        & $\boldsymbol{\pi_{0.5}}$
        & \textbf{UniVLA}
        & \textbf{OpenVLA} \\
        \midrule

        Spatial
        & 99/100 (99.0)
        & 94/100 (94.0)
        & 77/100 (77.0) \\

        Object
        & 99/100 (99.0)
        & 96/100 (96.0)
        & 86/100 (86.0) \\

        Goal
        & 79/80 (98.8)
        & 72/80 (90.0)
        & 57/80 (71.3) \\

        LIBERO-10
        & 93/100 (93.0)
        & 83/100 (83.0)
        & 50/100 (50.0) \\

        \midrule

        \textbf{Overall}
        & \textbf{370/380 (97.4)}
        & \textbf{345/380 (90.8)}
        & \textbf{270/380 (71.1)} \\

        \bottomrule
    \end{tabular}
\end{table}

The same ranking is observed across all four suites:
$\pi_{0.5} >$ UniVLA $>$ OpenVLA.
Overall, $\pi_{0.5}$ exceeds UniVLA by 6.6 percentage points,
UniVLA exceeds OpenVLA by 19.7 points, and
$\pi_{0.5}$ exceeds OpenVLA by 26.3 points.
The largest separation occurs on LIBERO-10, where success decreases from
93.0\% to 83.0\% and 50.0\%.

These scores show \emph{how much} the policies differ, but not
\emph{where} those differences arise during physical interaction.

% ============================================================
\subsection{Interaction Capability Profiles}
\label{sec:interaction_profile}
% ============================================================

We next examine whether higher task success corresponds to uniformly stronger
interaction capability or to improvements at particular stages.
Table~\ref{tab:main_pacebench} reports the capability profile of each policy
and the corresponding pairwise differences.

\begin{table*}[t]
    \centering
    \caption{
        \textbf{Interaction capability profiles and cross-model differences.}
        Panel A reports successful / eligible records, capability rate (\%),
        and 95\% confidence interval.
        Panel B reports pairwise capability differences in percentage points.
        Bold values indicate statistically significant paired differences after
        multiple-comparison correction ($\alpha=0.05$).
        Stable Hold is evaluated over channels in which a grasp event occurs,
        including grasps acquired beyond the 40-step Grasp window.
        Restricting Stable Hold to channels with a successful Grasp stage changes
        the reported rates by at most 0.2 percentage points.
    }
    \label{tab:main_pacebench}

    \renewcommand{\arraystretch}{1.10}
    \footnotesize

    \textbf{Panel A: Interaction Capability Profile}

    \vspace{1mm}

    \setlength{\tabcolsep}{8pt}

    \begin{tabular}{lccc}
        \toprule
        \textbf{Capability}
        & $\boldsymbol{\pi_{0.5}}$
        & \textbf{UniVLA}
        & \textbf{OpenVLA} \\
        \midrule

        Approach
        & \makecell{438/440 (99.5)\\{\scriptsize [98.9,100.0]}}
        & \makecell{437/440 (99.3)\\{\scriptsize [98.4,100.0]}}
        & \makecell{412/440 (93.6)\\{\scriptsize [91.1,96.0]}} \\

        Contact
        & \makecell{429/438 (97.9)\\{\scriptsize [96.6,99.1]}}
        & \makecell{428/437 (97.9)\\{\scriptsize [96.6,99.1]}}
        & \makecell{384/412 (93.2)\\{\scriptsize [90.8,95.5]}} \\

        Grasp
        & \makecell{328/349 (94.0)\\{\scriptsize [91.4,96.4]}}
        & \makecell{313/344 (91.0)\\{\scriptsize [87.8,93.9]}}
        & \makecell{271/320 (84.7)\\{\scriptsize [80.6,88.6]}} \\

        Stable Hold (given grasp)
        & \makecell{339/340 (99.7)\\{\scriptsize [99.1,100.0]}}
        & \makecell{325/330 (98.5)\\{\scriptsize [97.0,99.7]}}
        & \makecell{285/286 (99.7)\\{\scriptsize [98.9,100.0]}} \\

        GCG$_{\mathrm{end}}$
        & \makecell{337/342 (98.5)\\{\scriptsize [96.6,99.4]}}
        & \makecell{313/331 (94.6)\\{\scriptsize [91.6,96.5]}}
        & \makecell{256/288 (88.9)\\{\scriptsize [84.7,92.0]}} \\

        \bottomrule
    \end{tabular}

    \vspace{2.5mm}

    \textbf{Panel B: Cross-Model Capability Differences}

    \vspace{1mm}

    \setlength{\tabcolsep}{7pt}

    \begin{tabular}{lccc}
        \toprule
        \textbf{Capability}
        & $\boldsymbol{\pi_{0.5}-\mathrm{UniVLA}}$
        & \textbf{UniVLA--OpenVLA}
        & $\boldsymbol{\pi_{0.5}-\mathrm{OpenVLA}}$ \\
        \midrule

        Approach
        & +0.2
        & \textbf{+5.7}
        & \textbf{+5.9} \\

        Contact
        & +0.0
        & \textbf{+4.7}
        & \textbf{+4.7} \\

        Grasp
        & +3.0
        & \textbf{+6.3}
        & \textbf{+9.3} \\

        Stable Hold (given grasp)
        & +1.2
        & $-1.2$
        & +0.1 \\

        GCG$_{\mathrm{end}}$
        & \textbf{+4.0}
        & \textbf{+5.7}
        & \textbf{+9.6} \\

        \bottomrule
    \end{tabular}

\end{table*}

\noindent Capability differences are not uniform.
Compared with OpenVLA, UniVLA shows higher capability in Approach
(+5.7 points), Contact (+4.7), Grasp (+6.3), and
GCG$_{\mathrm{end}}$ (+5.7).
The larger $\pi_{0.5}$--OpenVLA gap shows the same broad pattern.

By contrast, $\pi_{0.5}$ and UniVLA are nearly identical in Approach and
Contact.
Their 3.0-point Grasp difference is not statistically significant
($p=.32$), while the clearest separation appears downstream at
GCG$_{\mathrm{end}}$, where $\pi_{0.5}$ exceeds UniVLA by 4.0 points
($p=.012$).

\noindent Conditional on an observed grasp, Stable Hold is already near
saturation.
Stable-Hold capability ranges from 98.5\% to 99.7\% across all three policies.
Restricting the analysis to channels with a successful Grasp stage yields
nearly identical rates of 99.7\%, 98.7\%, and 99.6\% for $\pi_{0.5}$,
UniVLA, and OpenVLA, respectively.
Notably, OpenVLA reaches 99.7\% despite trailing $\pi_{0.5}$ by
26.3 percentage points in overall task success.
Thus, higher task success does not imply uniformly higher capability across
all interaction stages.

\noindent\textbf{Sensitivity to evaluator parameters.}
The capability profiles are robust to moderate variations of the evaluator
parameters.
Varying the predicate activation thresholds by $\pm1$ timestep or increasing
the deactivation thresholds by two timesteps changes individual capability
rates by at most 1.2 percentage points.
The Grasp acquisition window has a larger effect on the absolute Grasp rate:
changing the window from the default 40 timesteps to 30 or 60 timesteps
changes Grasp capability as expected, but preserves the cross-policy ordering
and similar pairwise gaps.
Across all tested settings, $\pi_{0.5} >$ UniVLA $>$ OpenVLA remains unchanged
for Approach, Contact, Grasp, and GCG$_{\mathrm{end}}$, while Stable Hold
remains near saturation (98.5--100\%).

% ============================================================
\subsection{Which Failures Account for the Performance Gap?}
\label{sec:matched_gap_results}

Capability profiles reveal where policies differ, but they do not determine
which differences compose the observed task-success gap.
We therefore apply the matched performance-gap attribution from
Sec.~\ref{sec:PACE_matched_gap}.
Table~\ref{tab:matched_gap} summarizes the resulting attribution across the
three policy pairs.

\begin{table*}[t]
    \centering
    \caption{
        \textbf{Matched performance-gap attribution.}
        $W/L$ denotes episodes solved only by the higher-/lower-performing
        policy. Net gain is $W-L$, measured in matched episodes.
        The remaining columns report the signed normalized contribution of
        each interaction region to the net task-success gap; all grouped
        contributions in the reported comparisons are positive.
    }
    \label{tab:matched_gap}

    \renewcommand{\arraystretch}{1.12}
    \setlength{\tabcolsep}{8pt}
    \footnotesize

    \begin{tabular}{lccccc}
        \toprule
        \textbf{Policy pair}
        & \textbf{$W/L$}
        & \textbf{Net gain (eps.)}
        & \textbf{Pre-grasp}
        & \textbf{Post-grasp}
        & \textbf{Unclass} \\
        \midrule

        $\pi_{0.5}$--UniVLA
        & 31/6
        & +25
        & 28\%
        & \textbf{40\%}
        & 32\% \\

        UniVLA--OpenVLA
        & 97/22
        & +75
        & \textbf{60\%}
        & 31\%
        & 9\% \\

        $\pi_{0.5}$--OpenVLA
        & 107/7
        & +100
        & \textbf{52\%}
        & 33\%
        & 15\% \\

        \bottomrule
    \end{tabular}

    \vspace{0.5mm}

    {\scriptsize
    Pre-grasp = Approach + Contact + Grasp;
    Post-grasp = Stable Hold + Release + Place/Insert.
    Bold indicates the largest classified contribution for each policy pair.
    }
\end{table*}

\noindent\textbf{Large gaps involving OpenVLA are dominated by pre-grasp
failures.}
The 19.7-point UniVLA--OpenVLA gap corresponds to a net advantage of
75 matched episodes.
Of this gap, 60\% is associated with pre-grasp failures
(95\% CI: 46--76\%), compared with 31\% for post-grasp execution
(16--44\%).
At the individual-stage level, Grasp contributes +20 net episodes,
Place/Insert +19, and Approach +17.
Thus, most of the UniVLA--OpenVLA gap is associated with failures occurring
before or during object acquisition.

The $\pi_{0.5}$--OpenVLA comparison shows a similar pattern:
52\% of the net gap is associated with pre-grasp failures
(95\% CI: 42--63\%), compared with 33\% for post-grasp execution
(23--43\%).

\noindent The remaining $\pi_{0.5}$--UniVLA gap is more downstream.
The smaller 6.6-point gap corresponds to a net advantage of only
25 matched episodes.
Pre-grasp stages account for 28\% of the gap
(95\% CI: 4--52\%), while post-grasp stages account for 40\%
(13--65\%).
Among individual classified stages, \textsc{Place/Insert} is the largest
contributor, accounting for +9 net episodes, or 36\% of the total net gap.

These results show that different policy gaps have distinct failure-mode
compositions.
The gaps involving OpenVLA are primarily associated with early interaction
failures, whereas the smaller $\pi_{0.5}$--UniVLA gap has its largest
classified contribution in downstream task execution.

The latter result should be interpreted cautiously.
The $\pi_{0.5}$--UniVLA net difference contains only 25 episodes, 32\% of
which remain Unclassifiable, and the confidence intervals for the pre- and
post-grasp contributions overlap substantially.
We therefore do not claim that post-grasp contribution is statistically
greater than pre-grasp contribution.

% ============================================================
\subsection{Overall Failure Composition}
\label{sec:failure_results}

The matched analysis characterizes which failures are associated with
cross-policy performance differences.
Table~\ref{tab:failure_attribution} additionally summarizes where unsuccessful
rollouts break down overall.

\begin{table}[t]
    \centering
    \caption{
        Stage-wise attribution of unsuccessful episodes.
        Values are count (\% of failed episodes).
    }
    \label{tab:failure_attribution}

    \setlength{\tabcolsep}{2.8pt}
    \renewcommand{\arraystretch}{1.05}
    \footnotesize

    \begin{tabular}{lccc}
        \toprule
        \textbf{Missing stage}
        & $\boldsymbol{\pi_{0.5}}$
        & \textbf{UniVLA}
        & \textbf{OpenVLA} \\
        \midrule

        Approach
        & 0 (0.0)
        & 1 (2.9)
        & 19 (17.3) \\

        Contact
        & 0 (0.0)
        & 3 (8.6)
        & 13 (11.8) \\

        Grasp
        & 3 (30.0)
        & 7 (20.0)
        & 24 (21.8) \\

        Stable Hold
        & 0 (0.0)
        & 1 (2.9)
        & 1 (0.9) \\

        Release
        & 0 (0.0)
        & 0 (0.0)
        & 4 (3.6) \\

        Place/Insert
        & 5 (50.0)
        & 14 (40.0)
        & 33 (30.0) \\

        Unclassifiable
        & 2 (20.0)
        & 9 (25.7)
        & 16 (14.5) \\

        \midrule

        \textbf{Total failures}
        & \textbf{10}
        & \textbf{35}
        & \textbf{110} \\

        \bottomrule
    \end{tabular}
\end{table}

Early interaction failures are much more common for OpenVLA.
Approach and Contact together account for 29.1\% of its failed episodes,
compared with 11.5\% for UniVLA and none for $\pi_{0.5}$.
For the two higher-performing policies, the remaining failures are more
frequently associated with Grasp and Place/Insert.

A non-negligible fraction remains Unclassifiable:
20.0\% for $\pi_{0.5}$, 25.7\% for UniVLA, and 14.5\% for OpenVLA.
We therefore interpret these labels as structured behavioral attribution rather
than an exhaustive causal taxonomy.

The multi-object aggregation rule is also insensitive to an alternative
definition based on the earliest missing stage across objects.
Replacing the default goal-order rule changes the assigned label for only
0/4, 1/9, and 4/29 multi-object failed episodes with an identifiable missing
stage for $\pi_{0.5}$, UniVLA, and OpenVLA, respectively.
% ============================================================
\section{Limitations and Future Work}
\label{sec:limitations}
% ============================================================

PACE-Bench currently has several limitations.
First, our evaluation is restricted to three VLA policies on LIBERO, and the
interaction ontology is derived from simulator-accessible state.
Its generalization to additional benchmarks and real-robot settings therefore
remains to be validated.

Second, the current interaction stages are intentionally compact and do not
capture every possible failure mode.
Accordingly, 14.5--25.7\% of failed episodes remain unclassifiable under the
present taxonomy.
The current stage attribution is rule-based and has not yet been independently
validated against human failure annotations; the resulting labels should
therefore be interpreted as evaluator-defined behavioral categories rather
than ground-truth failure causes.
Because capability metrics use stage-specific eligible sets that can differ
across policies, cross-model capability differences should also be interpreted
as comparisons of conditional stage rates rather than unconditional
capabilities over a common task-defined population.
Some capabilities are also close to saturation; in particular, conditional
on an observed grasp, Stable Hold reaches 98.5--99.7\% across all evaluated
policies and therefore provides limited discriminative power in the current
setting.

Third, GCG$_{\mathrm{end}}$ is designed to localize post-grasp task completion
rather than serve as an independent replacement for task success.
Likewise, naturally occurring recovery events are too sparse to support strong
cross-policy recovery conclusions.

Future work will extend PACE-Bench to broader manipulation suites and
real-robot evaluation, enrich the interaction ontology with additional
task-dependent stages, validate the failure taxonomy against human judgments,
and investigate whether capability-level feedback can be used not only for
evaluation but also to guide data collection and targeted policy improvement.
% ============================================================
\section{Conclusion}
\label{sec:conclusion}
% ============================================================

We introduced PACE-Bench, an interaction-centric evaluation framework for
Vision-Language-Action policies that goes beyond ranking models by aggregate
task success.
PACE-Bench maps ordinary closed-loop execution into physically meaningful
interaction stages and, through matched episodes, behaviorally decomposes an
observed task-success difference into stage-specific failure contributions.

Across $\pi_{0.5}$, UniVLA, and OpenVLA on four LIBERO suites, we find that
performance gaps have different interaction structures.
For UniVLA over OpenVLA, 60\% of the net success gap is associated with
pre-grasp failures, indicating that much of the separation occurs before or
during object acquisition.
In contrast, for the smaller $\pi_{0.5}$--UniVLA gap,
\textsc{Place/Insert} is the largest classified contributor, accounting for
36\% of the net difference.
Conditional on an observed grasp, Stable Hold is already near saturation
across all three policies (98.5--99.7\%) despite task-success differences of
up to 26.3 percentage points.

These results show that higher task success should not be interpreted as a
uniform improvement across manipulation capabilities.
PACE-Bench turns an aggregate benchmark difference into a stage-level
behavioral diagnosis, revealing which interaction stages distinguish policies
and which capabilities are already saturated.
This attribution can support more targeted model design, data collection, and
mechanism validation in future VLA research.
\bibliographystyle{IEEEtran}
\bibliography{references}

\end{document}